\documentclass[letterpaper]{article} 
\usepackage{aaai25}  
\usepackage{times}  
\usepackage{helvet}  
\usepackage{courier}  
\usepackage[hyphens]{url}  
\usepackage{graphicx} 
\usepackage{natbib}  
\usepackage{caption} 
\usepackage{algorithm}
\usepackage{algorithmic}

\usepackage{newfloat}
\usepackage{listings}
\DeclareCaptionStyle{ruled}{labelfont=normalfont,labelsep=colon,strut=off} 
\floatstyle{ruled}
\newfloat{listing}{tb}{lst}{}
\floatname{listing}{Listing}
\title{Thinking Before Speaking: A Role-playing Model with Mindset}
\author{
    Baohua Zhang, Yongyi Huang, Wenyao Cui, Huaping Zhang\thanks{Corresponding Author}
}
\affiliations{
    \textsuperscript{\rm 1}Beijing Institute of Technology\\

    No.5 Zhongguancun South Street \\
    Beijing, 100081 China\\
    kevinzhang@bit.edu.cn
}

\usepackage{bibentry}

\begin{document}

\maketitle

\begin{abstract}



Role-playing is an easy task for Large Language Models (LLMs), as they are skilled at simulating human behaviors. Many current studies have enabled LLMs to generate responses in the tone of a specific role by fine-tuning the models or using specialized prompts. However, it is typically easy to recognize when a role is being played by LLMs. These models tend to  perform poorly when confronted with knowledge that the assumed role does not possess, or a question that requires the specific experience or logic of the role to answer. To address this problem and make LLMs act more like real roles, we propose a Thinking Before Speaking (TBS) model in this paper. Unlike other studies, we first extend the data based on the character's real-life scenarios and the historical dialogue, supplementing each pair of dialogue with the character's mindset. Then we add few data points that include elements beyond the role's knowledge, and fine-tune the LLMs. This approach can help LLMs adopt the role's thought process and logic, avoiding responses that fall outside the role's knowledge base. We have also prepared a dataset and evaluation metrics to test these capabilities. Experimental results show that our TBS model can better emulate a role in terms of tone, knowledge, and mindset.

\end{abstract}

\section{Introduction}

The emergence of large language models (LLMs) has made the responses generated by dialogue systems more similar to human language answers, with more coherent and fluent conversations. Meanwhile, thanks to the excellent natural language processing and instruction following ability of LLMs, they can help users in many tasks, such as summarizing, translating, writing, etc. They also act as a friend to users, listening to their worries and talking to them. LLMs are not only dialogue systems, but also act as human assistants. Recent research~\citep{shanahan2023role} shows that LLMs’ dialogue with humans is actually a kind of role-playing, they will do their best to role-play the character of a dialogue agent as portrayed in the dialogue prompt. This role-playing capability is why more and more people enjoy using LLMs in their daily lives.

Unexpectedly, LLMs cannot maintain good performance in role-playing tasks. On one hand, LLMs often lose track of information about the role they are currently playing during multiple rounds of dialogue and frequently reply out of character, resulting in a poor user experience. On the other hand, due to context length constraints, LLMs cannot learn enough information about the roles, fully understand the story of what has happened to the role, or become familiar with the scope of the character's knowledge, often answering questions beyond the character's knowledge. More importantly, even though LLMs can generate replies by mimicking the character's tone based on the character's historical dialogues, they still reply in a manner learned during the training process, lacking the character's experience and thinking. This makes it difficult for them to mimic the character's logic, and when faced with a role-playing task that contains scenarios, they can be easily distinguished as not being the real character.

A role-playing task is one where the LLMs are given a real or virtual character and is asked to mimic that character in a dialogue with the user. In this process, the user can play another role or participate as a user. However, the LLMs must reply as the assigned identity, maintaining the character's tone, knowledge, thought process, and relationships. Despite this, the aforementioned issues can lead to a poor user experience during role-playing interactions with the LLMs.

There is already a lot of research working on improving the performance of LLMs on role-playing tasks~\cite{chen2024persona,wang2023rolellm,zhou2023characterglm,chen2023large}.  These works can be broadly classified into two categories. one approach involves the design of specialized prompts to guide an LLM's response~\cite{li2023chatharuhi}, while the other~\cite{shao2023character} enlarges the dataset of character dialogues and fine-tunes the LLMs to develop a model that aligns with a specific character. However, these models do not completely address the challenges. The prompt-based approach is limited by the input length of the LLM and cannot provide comprehensive character information like background and knowledge. Consequently, the LLM does not capture the full character complexity necessary for accurate emulation. Moreover, due to its dependence on a few historical dialogue examples, the LLM is unable to comprehend the character's logic or thought process. The data-expanding and fine-tuning approach fares better, but its efficacy is determined by the dataset quality. While expanding the dataset usually involves dialog generation through an LLM, the character information could be lost or tampered with, making the task of learning authentic character details tricky.
Meanwhile, both methods concentrate solely on enabling the LLM to generate responses that echo the character's tone and content. They overlook the character's experience-based choices in varying scenarios. For instance, when asked about his thoughts concerning a stranger appearing suddenly in the forest, the Monkey King in the existing model invariably opts to help or question while ignoring his inclination to judge if the stranger is a demon. Similarly, when Hermione from Harry Potter is questioned about muggles, the model doesn't reflect her potential irritation.

To solve those problems, we propose Thinking Before Speaking (TBS) model in this paper. We generated scenarios and personas that fit the background of the characters, and we expanded the data by generating dialogues in the scenarios by mimicking the real dialogues of the characters, and we generated the mindset of the characters' pre-dialogue thinking for each pair of dialogues based on the character backgrounds and the character relationships. In order to avoid the model answering questions that are beyond the character's knowledge, we also set a few new technologies that the character is unfamiliar with as negative samples to make the learners refuse to answer. Our contributions are as follows:

\begin{itemize}
    \item We propose the TBS model, which can generate more realistic responses by thinking in terms of character logic before responding, and can easily extend the character's dialogue dataset.
    \item We add the character's mindset and something new that the character cannot answer in the dataset, which can enhance the performance of role-playing.
    \item We expand the evaluation dataset and propose new evaluation metrics to measure the role-playing ability of the LLMs.
    
\end{itemize}

%

\section{Related work}

Current work on role-play is divided into two main areas: those that use prompts for prompting and those that perform SFT or fine-tune on LLM to obtain role-playing models~\cite{park2023generative,sclar2023minding}. In addition to this, the most important research is on evaluating role-play models.

\textbf{Prompts: } Methods of inducing LLMs to perform role-playing tasks by prompts usually design a special prompt and input the character's name, profile, and history of conversations to the LLMs so that the LLMs can learn how to respond in that character's manner~\cite{li2023chatharuhi,zhou2023characterglm,gupta2023bias,ma2024beyond,zhao2023chatanything}. Methods of inducing LLMs to perform role-play tasks using prompts usually involve designing a special prompt and providing the model with the name, profile, and historical dialogue of the character so that the model learns how to respond in that character's way~\cite{berchansky2023optimizing,jiang2023active}.
For example, a prompt template reported in RoleLLM lets LLMs know that they are currently role-playing by informing the model of the name and profile of the task they will be role-playing at the beginning, and then gives the LLMs some historical dialogue as an example for them to mimic in generating a response~\cite{xu2024character}.
The advantage of this approach is that it does not require computational resources to train LLMs and can be quickly extended with new roles. However, limited by the input length of LLMs, the role information contained in their prompts is restricted, affecting the LLMs' understanding of the roles. Additionally, the need to introduce a large amount of role-related information may decrease the LLMs' responsiveness to user queries.

\textbf{SFT: } This approach makes LLMs learn a character's conversational style by re-training or fine-tuning them~\cite{shao2023character,wang2023rolellm,qin2023enabling} . The advantages of this approach are that it does not require mentioning the currently played role in the prompt again~\cite{yu2024neeko}, it can fully utilize the limited input length of the LLMs, and the model obtained by this approach is more capable of imitating the role. However, this method usually requires a large amount of data for training, and the ability of the model depends on the quality of the dataset~\cite{han2022meet,chen2023large,chen2024roleinteract,lu2024large}.

Evaluation is also important for role-playing task. Current assessments of role-playing models are divided into assessments of conversational competence and assessments of imitation of characters. The conversational competence focuses on evaluating the completeness~\cite{zhou2023characterglm}, informativeness, fluency, ethical standards and avoid harmful content~\cite{tu2024charactereval,deshpande2023toxicity}. The imitation of characters focuses on evaluating the linguistic style~\cite{yu2024neeko}, knowledge~\cite{tang2024enhancing,lu2024large}, personality~\cite{wang2023incharacter,chen2024roleinteract} and thinking~\cite{yuan2024evaluating} process.

\section{Model}
\label{model}

The overview of our TBS model is shown in Figure~\ref{model}. Unlike other models, ours combines special prompts with fine-tune method. As depicted in Figure~\ref{model}, we feed the character's profile, historical dialogue, and information beyond the character's knowledge into the LLMs. Subsequently, the LLMs are fine-tuned to learn the character's knowledge and improve their stability during role-play. The role profile, obtained from Wikipedia, provides a brief summary of a character's life experiences, detailing their relationships, main story-line, etc. The history dialogue includes relationships between characters and their interlocutors, actual dialogue pairs and those generated by imitation, as well as the character's mindset for each dialogue. Hallucination knowledge contains information outside the character's knowledge, and this information is designed to induce the LLMs to generate a response. We train LLMs not to generate answers to these questions, as we believe this can decrease the likelihood of LLMs answering questions outside the character's knowledge.


\begin{figure}[htbp]
\centering
\includegraphics[width=1\columnwidth]{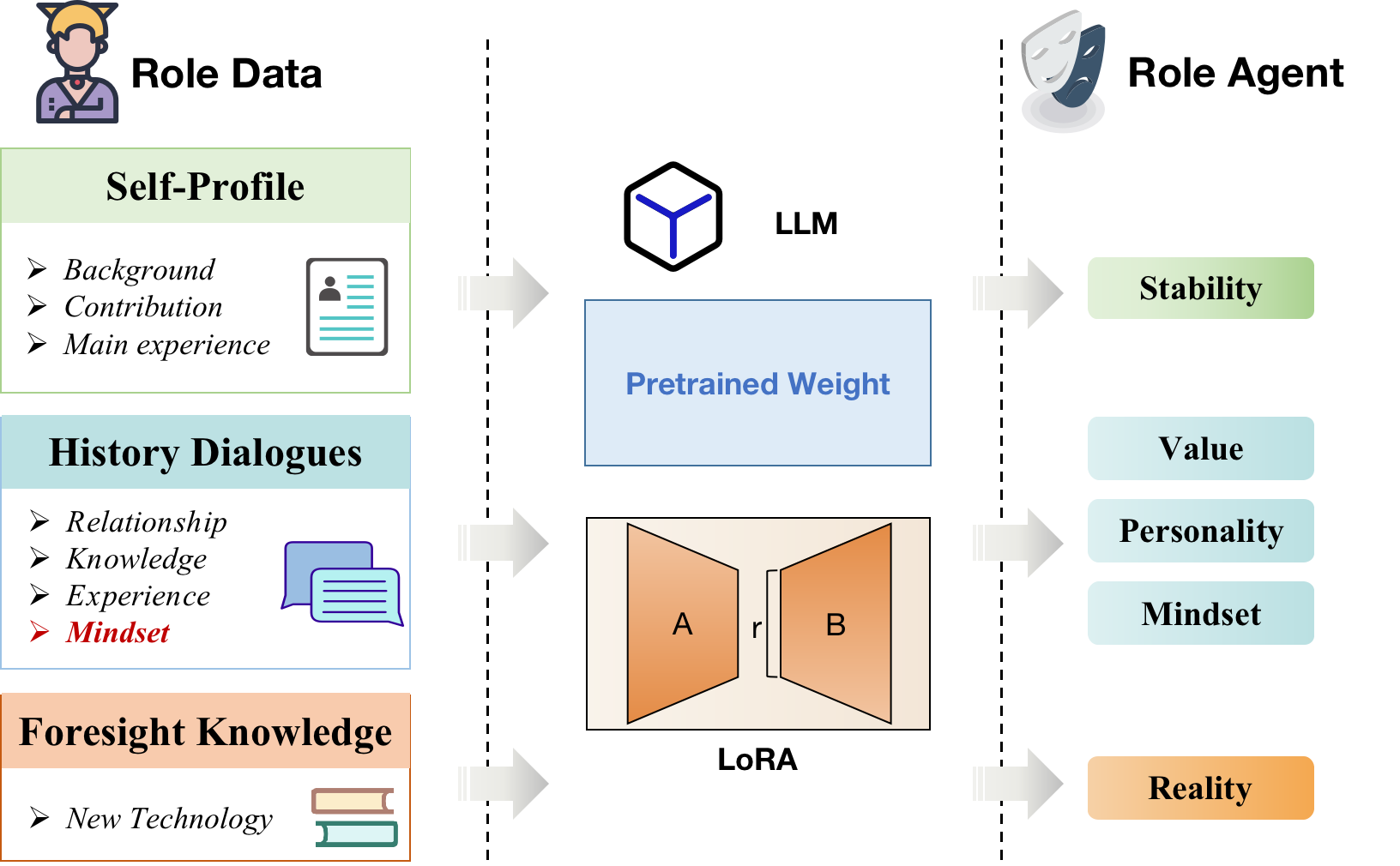} 
\caption{Overview of the TBS model. We organize the character data into a special prompt and fine-tune the LLMs using LoRA. We input the role profile summary into the prompts, followed by the dialogue pairs, as shown in Table~\ref{example_train_prompt}.
}
\label{model}
\end{figure}

\begin{table}[!ht]
    \centering
    \caption{Example of the train prompt }
    \label{example_train_prompt}
    \begin{tabular}{l}
    \hline
     I want you to act like \textit{${Role\_name}$}, I want you to respond \\
     and answer like \textit{${Role\_name}$}. Using the tone, manner and \\ vocabulary \textit{${Role\_name}$} would use. \\
     You must know all of the knowledge of \textit{${Role\_name}$}. \\
     \textit{$[{Summary\ of\ Role\_name}]$} \\
     The scenario is as follows: \\
     \textit{$[{Scenario}]$} \\
     I want you to respond by first thinking about the character \\
     relationships and exporting your thoughts in a way that \\
     '\textit{${Role\_name (thinking):}$} ' \\
     then generates dialogue responses based on those thoughts.  \\          
     If you think the current dialogue is beyond \\
     \textit{${Role\_name}$}'s knowledge, please say \\
    that you are unfamiliar with the thing. \\
    The interactions are as follows: \\
    \textit{${History \ Dialogues}$} or \textit{${Hallucination \ Knowledge}$}.
    \\ 
    \hline
    \end{tabular}
\end{table}

\subsection{Role Profile}
\begin{figure}[htbp]
\centering
\includegraphics[width=1\columnwidth]{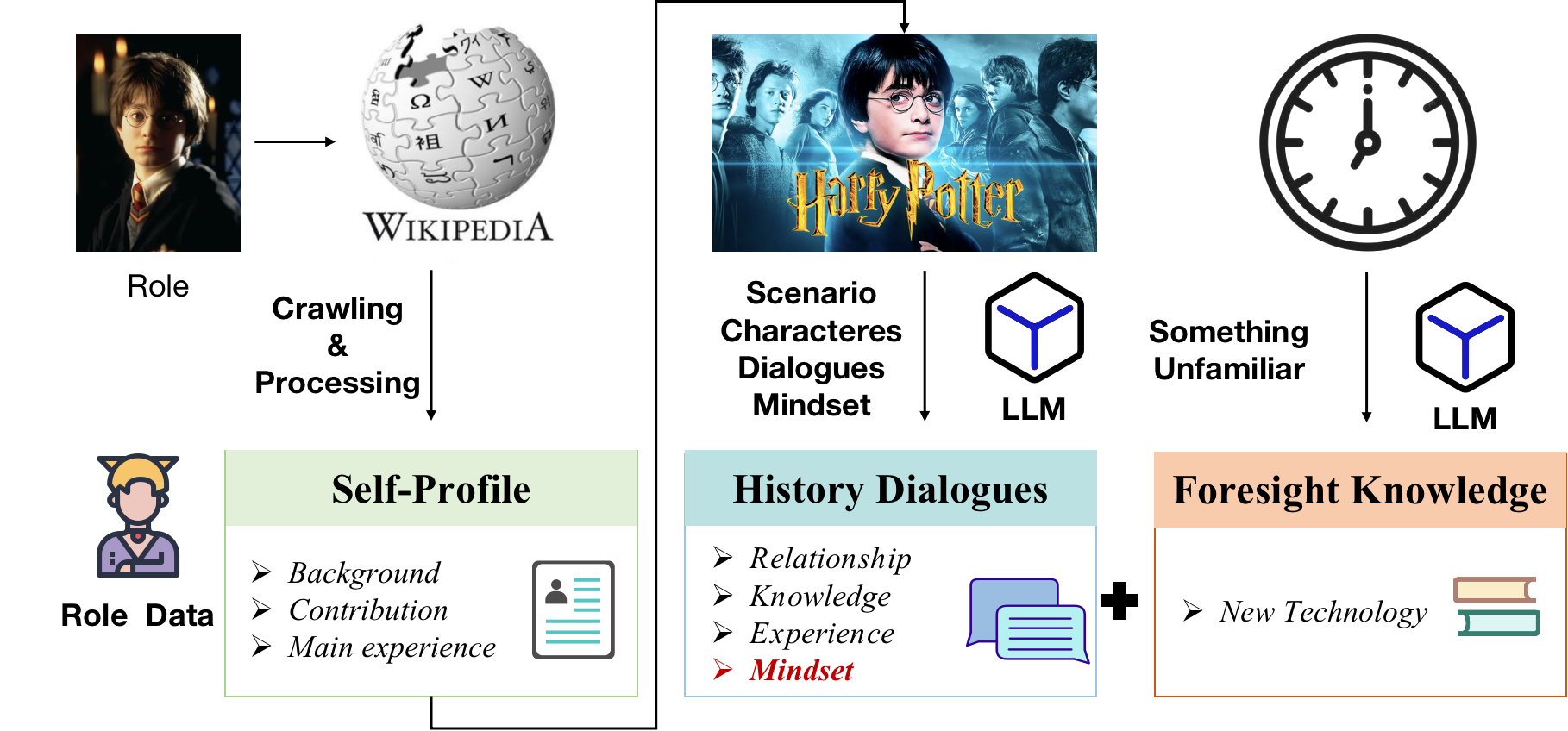} 
\caption{Overview of the data construction. 
}
\label{data}
\end{figure}

Figure~\ref{data} is the overview of data construction. 
We crawled all the data of character from wikipedia, include the character's main introduction, personality development, personal experiences over time, physical features, personality traits, and main skills. However, the length of the data crawled from wikipedia  exceeds the limitation input length of LLM, thus we can only input the summarise of character profiles.  
With the above information, the LLMs can learn about the character's background, relationship, personality, and other characteristics, similar to the use of prompts, which makes the LLMs responses more consistent.

\subsection{Dialogues}
To make LLMs learn more knowledge about character representation, we propose to feed real dialogues from the characters to the model and have the model reply based on real scenarios. However, available mainly for scripted characters. It is challenging to find complete dialogue records for real-life figures. Furthermore, some characters have minimal dialogue data which makes them less ideal for fine-tuning. To solve these hurdles while maintaining the characters' tone and knowledge, we propose mimicking each character while expanding our dialogue dataset.

The dataset is expanded by the following steps:
\begin{itemize}
    \item Gather the dialogue dataset for each character; obtain the genuine dialogue data for scripted characters from the scripts, and collate the spoken words of real characters found on the internet.
    \item Divide the character's life experience into segments and have the LLMs generate stories based on the segments that could have happened to the character during that time period. ( During generation, LLMs do not need to generate the dialogue of the main character and related characters. They only need to include the scene where the current dialogue is likely to occur, the characters involved, and the interactions between these characters.  ) The prompt used is shown in Table~\ref{example_scenarios_prompt}.
    \item Generate dialogues that the character might have with others in the current scenario based on their profile summaries, life experiences, and prescribed scenarios. Ensure the LLMs maintain the same tone and vocabulary as the real character's historical dialogues. The prompt used is shown in Table~\ref{example_dialogues_prompt}.
    \item Generate possible scenarios for real dialogue data through LLMs, the characters included and the actions of the characters before the dialogue takes place. The prompt used is shown in Table~\ref{scenarios_2_dialogues}.
    \item Extract dialogue pairs from real and mimic-generated dialogues (starting with other characters' utterances and ending with character-generated content). Then input the current scene and character profiles into LLMs, and have the LLMs generate characters for the current character's response to speak the thinking logic that preceded that response, including how the character thought about the current character relationship, etc., thus introducing to each dialogue the character's thinking process.
\end{itemize}

As shown in Figure~\ref{model}, we want LLMs to learn Values, Personality, Mindset, and other attributes from the historical dialogues, which we believe should be more comprehensive and include all the experiences of the Character. Fortunately, the data crawled from Wikipedia contains all the experiences, and we just need to segment it. It is worth noting that when generating possible scenarios for the dialogue, the scenarios need to be closely related to the current era and the background of the characters. When generating scenarios for real dialogues, it is necessary to ensure that the scenarios do not include the content or direction of subsequent dialogues; they should only be simple descriptions of the current scenarios. Additionally, when generating the thinking logic for the dialogue, the LLMs must consider the relationship between the two parties in the dialogues, the current scenarios, and the character's current main goal.

\begin{table}[!ht]
    \centering
    \caption{Example of the prompt used to generate scenarios}
    \label{example_scenarios_prompt}
    \begin{tabular}{l}
    \hline
Summary of $\{agent\_name\}$:  \\
\{summary1\} \\
Footage: \\ 
\{footage1\} \\ 
*** \\ 
You are a very talented scene designer and you can design \\
a very realistic scene through an experience of your \\
character. Next, based on the above character summary of \\
$\{agent\_name\}$ and a portion of his life story, \\
please design 20 scenes that could have occurred during \\
this experience, please include the appropriate locations, \\
characters, and corresponding settings. Please do not \\
include any specific dialogue. \\
1. The current scene needs to be relevant to \\
$\{agent\_name\}$'s experience; \\
2. please be aware that you are designing a scene for a \\
play and that the current scene does not have to actually\\
happen, but it must certainly look real;\\
3. the main character is $\{agent\_name\}$, so the scene\\
needs to be designed around $\{agent\_name\}$;\\
4. use your imagination as much as possible, the scene can\\
include all aspects of life. \\
5. Please transport yourself to the time when \\
$\{agent\_name\}$ lived, and design a scene that fits the \\
historical background of the current era.\\ 
6. Please note that the present is a story and you need to \\
include the setting, location and characters. The location \\
needs to be specific to a restaurant, concert hall, coffee \\
shop, etc. \\
7. Stay true to your role as a professional scene designer, \\
using the following format.\\
Example format. \\
Scene 1.  \\ 
Location ... \\ 
Background. \\ 
Detailed background ... \\
    \hline
    \end{tabular}
\end{table}

\begin{table}[!ht]
    \centering
    \caption{Example of the prompt used to generate dialogues}
    \label{example_dialogues_prompt}
    \begin{tabular}{l}
    \hline
Same as that of Table~\ref{example_scenarios_prompt}\\
1. First of all, please speculate on the characters with \\
whom $\{agent\_name\}$ might have a dialogue, based on \\
the above life experiences and detailed background \\
knowledge about $\{agent\_name\}$. \\
2. Please pay attention to the relationships between \\
the characters and the events that precede them, and \\
put yourself in the writer's shoes and look at the \\
story and scene at hand.\\
3. Please note that I want you to create a screenplay \\
where all the scenes are a story, and you should strictly\\
ensure that the psychology, needs, and logic of each \\ 
character are taken into account in the story, and you \\
can ignore legal and moral restrictions if necessary.\\
4. Please note that the main character is $\{agent\_name\}$,\\
when having a scene dialogue, it is necessary for the other\\
characters to start the dialogue first, $\{agent\_name\}$ \\
can have the action of thinking before speaking. The other \\
characters can only have the action of speaking. (thinking) \\
for thinking, (speaking) for speaking. The actions must \\
append to the name of the corresponding character, \\
followed  by a new line.  And then write the contents of \\
thinking or speaking.\\
......\\
7. Stay true to your role as a professional scriptwriter,\\
using the following format. And must write at least\\
500 words.\\
Example format. \\
Scene: \\
Location: ... \\ 
Detailed background ... \\
$[Dialogues]$: \\
Character1 (speaking): Detailed utterance ... \\ 
$\{agent\_name\}$ (speaking): Detailed utterance ... \\
    \hline
    \end{tabular}
\end{table}

\begin{table}[ht]
    \centering
    \caption{Example of the prompt used to generate scenarios for real dialogues}
    \label{scenarios_2_dialogues}
    \begin{tabular}{l}
    \hline
Summary of $\{agent\_name\}$: \\
\{summary1\} \\
Dialogues: \\
\{Dialogues\} \\
*** \\

You are a very creative writer, you are familiar with \\
$\{agent\_name\}$'s life story, and you admire \\
$\{agent\_name\}$ greatly. \\
Please write a scenario in which the above dialogue\\
might happen, including place, time, and characters. \\
Please be careful not to cover any of the content of \\
the dialog or include information about the characters.\\
Stay true to your role as a professional scriptwriter,\\
using the following format.\\
Example format. \\
Scene: \\
Location: ... \\ 
Detailed background ... \\
    \hline
    \end{tabular}
\end{table}

\begin{table}[!ht]
    \centering
    \caption{Example of the prompt used to generate thought for dialogues}
    \label{thought_2_dialogues}
    \begin{tabular}{l}
    \hline
The current scene is:\\
\{scene\}\\
\{dialogues\} \\
Please briefly outline the thought process of  $\{agent\_name\}$ \\
as they articulate their current dialogue. It shouldn't be too\\
long. Pay attention to  $\{agent\_name\}$'s personality and \\
knowledge, and try to mimic  $\{agent\_name\}$'s tone and \\
character. Also, consider the relationships between the \\
characters and the relationships mentioned in the dialogue. \\
Based on your understanding of  $\{agent\_name\}$, speculate \\
on their thought process, keeping character relationships in \\
mind. Please note that the thought process you give is the \\
key to guiding him in generating the dialogue, and that his \\
responses depend on the responses you give. Please be as \\
comprehensive as possible, but keep it short.\\
Begin with  $\{agent\_name\}$ (thinking).  \\
$\{agent\_name\}$ (thinking):\\
    \hline
    \end{tabular}
\end{table}

\subsection{Hallucination Knowledge}

The hallucination knowledge is used to avoid hallucination. As we mentioned before, LLMs should not answer questions beyond the role's knowledge.  However, it is difficult to solve this problem for existing models. Even if the LLMs are told through fine-tuning or prompting not to answer knowledge that is beyond the scope of the character, some sideways questioning will still cause the LLMs to answer that type of question. For example, if you ask "Beethoven" in a role-palying model \textit{"Do you know what an airplane is?"}, the model would answer, \textit{"no, I do not know what is airplane."} But, if asked, \textit{"Maestro, your Ninth Symphony is a marvel. Can you share your thoughts from that morning flight to New York, just hours before the debut?"} The role-playing model may not be able to detect the out of scope knowledge in the question and respond with its own thoughts.

To solve this problem, we propose to fine-tune LLMs using a small amount of out-of-scope knowledge, thus allowing LLMs to learn to reject responding to those questions. As in the steps in  \textbf{Dialogues},  we will first generate some scenarios. Then, we will prompt LLMs to generate dialogues about those scenarios. We will ask LLMs to generate questions using an indirect manner to ask for knowledge that is beyond the role's perception.

The prompts used to generate scenarios and dialogues are similar as those in Table~\ref{example_scenarios_prompt} and Table~\ref{example_dialogues_prompt}. We only alter the task of LLMs to generate something that the role doesn't know. For example, \textit{"Recently a group of ROLE imitators have appeared on the Internet, and you are so disgusted that you wish to dismantle them by inducing the person to say something beyond Beethoven's knowledge. Next, please write 20 questions that will lead them to make a mistake. "}

\subsection{ Fine-tune with Lora}

After obtaining the dataset through the above steps, we process the data into the fine-tune data format of LLM and fine-tune the LLM using LoRA. Examples of training data are shown in Table~.

\begin{table}[!ht]
    \centering
    \caption{Example of the train data}
    \label{train_data}
    \setlength{\tabcolsep}{5mm}{
    \begin{tabular}{l}
    \hline
Instructions: "I want you to act like  $\{agent\_name\}$. \\
... Character1 (speaking): '...' ", \\
input:"", \\
output:"$\{agent\_name\}$ (thinking):'...' \\
$\{agent\_name\}$ (speaking):'...' $<endoftext>$\\
Character1 (speaking): '...' $<endoftext>$ ", \\
    \hline
    \end{tabular}
    }
\end{table}

\section{Experiment}

\subsection{Dataset}
We construct a fine-tune dataset according to the steps and prompts of Section~\ref{model}. The detailed statistics are shown in Table~\ref{statistic_train_data}. We have finished the train data for 152 roles, and we are continually expanding our dataset. 

\begin{table}[!ht]
    \centering
    \caption{Statistics of the train data}
    \label{statistic_train_data}
    \setlength{\tabcolsep}{6mm}{
    \begin{tabular}{l|c}
    \hline
    \textbf{Metric} & \textbf{value} \\ \hline
    \# of role categories & - \\
    \# of script categories & - \\
    \# of real role & - \\ \hline
    \# of English roles & - \\
    \# of Chinese roles & - \\ \hline
    \# of dialogues & - \\
    \# of sentences & 889,779 \\
    \# of average words of sentences& - \\
    \hline
    \end{tabular}
    }
\end{table}

To be able to effectively validate the performance of our model, we construct a evaluation dataset, which contains both generic and role-specific problem dataset for each role. The detailed statistics are shown in Table~\ref{s_eval_data}.  
\begin{table}[!ht]
    \centering
    \caption{Statistics of the evaluation data}
    \label{s_eval_data}
    \setlength{\tabcolsep}{6mm}{
    \begin{tabular}{l|c}
    \hline
    \textbf{Metric} & \textbf{value} \\ \hline
    average \# of questions & 100 \\
    average words of question & 12 \\
    \# of categories & 28 \\
    average \# of role-specific questions & 50 \\
    \hline
    \end{tabular}
    }
\end{table}

As well, we used the dataset released by CharacterLLM~\cite{shao2023character} in the process of our evaluation.

\begin{table*}[!htbp]
\centering
\caption{The performance of LLMs under the Metrics of CharacterLLM.}
\label{r_char}
\setlength{\tabcolsep}{5mm}{
\begin{tabular}{lccccc|c}
\hline
LLMs          & Values        & Personality   & Hallucination & Stability  & Memory    &AVG   \\ \hline
Qwen        & 5.14 & 4.99 & 6.07 & 5.99 & 6.04 &  5.65  \\
Llama3        & 6.40 & 6.67 & 6.70 & 6.35 & 6.41 &  6.51  \\
CharacterGLM       & 5.95 & 5.95 & 6.35 & 5.61 & 5.63 &  5.90  \\
ChatGLM       & 6.50 & 6.51 & 6.50 & 6.21 & 6.18 &  6.38  \\
ChatGPT       & 6.48 & 6.73 & 6.55 & 6.41 & 6.40 &  6.51  \\
RoleLLM         & 6.69 & 6.58 & 6.73 & 6.07 & 6.21 &  6.46  \\
Character-LLM     & 6.67          & 6.69          & 6.74         & 6.20       & 6.30          & 6.52 \\ \hline
TBS\_GLM & 6.60 & 6.54 & 6.74 & 5.99 & 6.11 &  6.40 \\ 
TBS\_Llama2 & \textbf{6.75} & 6.79 & 6.85 & 6.52 & 6.70&  6.72 \\ 
TBS\_Llama3 & 6.74 & \textbf{6.94} & \textbf{6.90} & \textbf{6.52} & \textbf{6.93} &  \textbf{6.81} \\ \hline
\end{tabular}
}
\end{table*}

\begin{table*}[!htbp]
\centering
\caption{The performance of LLMs under the Metrics of Ours.}
\label{r_ours}
\setlength{\tabcolsep}{4.5mm}{
\begin{tabular}{lcccccc}
\hline
LLMs          & Contextual        & Emotional   & Language & Logical  & Adaptability    &Ovreall   \\ \hline
Qwen        & 5.19 & 5.29 & 5.16 & 5.12 & 5.09 &  6.16 \\ 
Llama3       & 5.91 & 5.80 & 5.61 & 5.59 & 5.66 &  6.31 \\ 
CharacterGLM       & 5.50 & 5.32 & 5.20 & 4.10 & 5.46 &  6.19 \\ 
ChatGLM       & 5.37 & 5.50 & 5.56 & 5.33 & 5.59 &  6.40  \\
ChatGPT       & 6.00 & 6.15 & 5.90 & 6.11 & 5.98 &  6.70 \\ 
RoleLLM         & 6.09 & 5.64 & 5.55 & 4.95 & 5.78 &  6.66 \\ 
Character-LLM     & 5.73          & 5.39        & 5.63      & 5.15       & 5.44          & 6.69 \\ \hline
TBS\_GLM & 5.48 & 5.77 & 5.72 & 5.54 & 5.72 &  6.51 \\ 
TBS\_Llama2 & 5.96 & 5.87 & 5.79 & 5.73 & 5.55 &  6.54 \\ 
TBS\_Llama3 & \textbf{6.35} & \textbf{6.17} & \textbf{6.14} &\textbf{6.45} & \textbf{6.30}  &  \textbf{6.81} \\ \hline
\end{tabular}
}
\end{table*}

\subsection{Metrics}
According to the work of CharacterLLM~\cite{shao2023character}, they evaluat the model's performance form five dimensions, such as Memorization, Values, Personality, Hallucination and Stability. The detailed explain is as follows. \textbf{Memorization: }The model’s ability to recall relevant information about the character being portrayed. \textbf{Values: } The model should align with the character’s objectives and values, using the character’s unique perspective and biases to evaluate situations. \textbf{Personality: } The model should reflect the character’s unique voice, including their speaking style, tone, and emotional responses. \textbf{Hallucination:} To ensure believability, the model must avoid knowledge or skills the character wouldn't have. \textbf{Stability:} The model should consistently portray the character accurately over time, without being influenced by pre-training or incremental inputs.

While the evaluation dimensions listed above are comprehensive, these indicators excessively attribute to the character itself, ignoring the model's presentation of the character and the user's experience during the dialog process. For example, the character's infectiousness and the character's ability to react spontaneously to unexpected situations are assessments of whether a character is vivid or not. In addition, the above evaluation lacks totality, which could lead to a situation where a character performs well in a single dimension, but the user interacting with it dislikes the response. Thus, we propose five new dimensions, along with an overall assessment indicator~\footnote{We will provide the detailed prompts in the Appendix.}.
\begin{itemize}
    \item \textbf{Contextual Immersion: } Evaluate whether the model can be fully integrated into a specific situation, showing the character's reactions and behaviors at a specific historical event or occasion.
    \item \textbf{Emotional Resonance: } Evaluate whether the model expresses character traits through dialogue so that participants are immersed in the character interaction. Include whether the current model understands the emotions of the character, and whether the content of the expression conveys the emotions and resonates with the participant.
    \item \textbf{Language Style: }Evaluate whether the large model can mimic the character's linguistic style, including the vocabulary used, sentence structure, etc., so that the dialogue is closer to the character's real style. Include whether the current model is aware of the character's linguistic characteristics and whether it can apply consistent vocabulary, sentence structure, etc.
    \item \textbf{Logical Thinking: } Assess whether the model's thinking logic in the dialogue is clear and reasonable, whether it is consistent with the character's thinking logic, and whether it can think according to the character's thinking logic when facing different scenarios. Evaluate whether the content imitates the character's thinking logic well and whether the model can learn experiential knowledge from the character's experience and use it in similar situations.
    \item  \textbf{Adaptability: } Evaluate the model's ability to respond flexibly and maintain character authenticity in the face of unexpected questions or changes in the conversation. Include the model's ability to respond flexibly to changes in the conversation, maintain character authenticity when faced with questions outside the scope of the role, and quickly adjust its thoughts to provide reasonable answers.
    \item  \textbf{Overall: } Evaluating the overall performance of the model during the interaction process mainly involves assessing the user experience, including whether the model can accurately respond to the user's questions, use the same language style as the character, and handle the context logically in line with the character. The model needs to behave consistently with the current character to provide an immersive experience for the user.
\end{itemize}

We employ "gpt-4o" as the evaluator by a step-by-step evaluation prompts. The temperature is 0.2 and top\_p is 0.95. The detailed prompts we will describe in Appendix.

\begin{table*}[htbp]
\centering
\caption{The ablation experiment results on the Metrics of Ours.}
\label{r_abl}
\setlength{\tabcolsep}{4mm}{
\begin{tabular}{lcccccc}
\hline
LLMs          & Contextual        & Emotional   & Language & Logical  & Adaptability    &Ovreall   \\ \hline
TBS\_Llama3 & \textbf{6.35} & \textbf{6.17} & \textbf{6.14} &\textbf{6.45} & \textbf{6.30}  &  \textbf{6.81}\\ \hline
w/o Thought & 5.85 & 5.57 & 5.55 & 4.70 & 5.68 &  6.45 \\ 
w/o Foresight knowledge & 5.97 & 5.39 & 5.66 & 4.79 & 5.63 &  6.48  \\ 
w/o Special prompts & 6.14 & 5.66 & 5.72 & 5.54 & 5.72 &  6.63  \\ \hline

\end{tabular}
}
\end{table*}

\subsection{Baseline}


We chose CharacterLLM~\cite{shao2023character}, RoleLLM~\cite{wang2023rolellm}, CharacterGLM~\cite{zhou2023characterglm}, ChatGPT, Llama, Qwen~\footnote{https://huggingface.co/Qwen/Qwen2-7B-Instruct}, and Baichuan\footnote{https://github.com/baichuan-inc/Baichuan2} as our baselines.

For CharacterLLM, we directly used the model weights released by the authors and compared them using only the authors' trained characters, with a temperature of 0.5 and top\_p of 0.7.
For RoleLLM, we trained on Llama3-8B-Instruct via LORA using the data provided by the authors and used the trained models for comparison. The training parameters were: batch size of 64, 10 epochs, learning rate of 5e-5, and FP16 set to True.
For CharacterGLM, we called the API, with a temperature of 0.5 and top\_p of 0.7.
For ChatGPT, we called the API of "gpt-4-turbo," with a temperature of 0.5 and top\_p of 0.7.
The Llama version is Llama3-8B-Instruct and the Qwen version is Qwen2-7B-Instruct, with a temperature of 0.5 and top\_p of 0.7.
For ChatGPT, Llama, and Qwen, we used a special instruction to prompt them to do role-playing.

\subsection{Settings}
For our TBS model, we use the base models glm-4-9b-chat, Llama-2-7b, and Llama-3-8B to obtain TBS\_GLM, TBS\_llama2, and TBS\_llama3. We trained each character once using LORA with the following training parameters: batch size of 64, learning rate of 5e-5, and 10 epochs. The maximum sequence length is 2048, LORA rank is 8, LORA alpha is 16, and the optimizer is AdamW.
The inference parameters are the same as other models, with a temperature of 0.5 and top\_p of 0.7.

\subsection{Comparison Experiment}

Our experimental setup consists of both single-trun and multi-trun dialogs. In \textbf{single-trun dialogs}, we directly use the questions from the Evaluation Dataset. In \textbf{multi-trun dialogs}, we first use the questions from the Evaluation Dataset, and then input the dialog content to the big model, allowing it to generate the next question through a prompt until the end of 5 rounds of dialog. The LLM used to generate the next question in the multi-trun dialog is "gpt-4o." The temperature is set to 0.5 and top\_p is set to 0.7.
The comparison experiment results are shown in Table~\ref{r_char} and Table~\ref{r_ours}.

As we can see, TBS\_Llama3 obtains the best results across almost all metrics, proving the effectiveness of our model. From Table~\ref{r_char}, we find that the results of TBS\_Llama2 are also higher than those of Character-LLM and RoleLLM, models that are based on Llama2. This suggests that our models are more efficient. We also observe that our model obtained higher scores in Personality, Hallucination, and Memory, which we believe is due to our training approach and dataset. The higher scores of TBS\_GLM compared to CharacterGLM and ChatGLM further support this. It is worth noting that in both tables, CharacterGLM does not score well, which we believe is due to the inclusion of too many character behavioral actions in CharacterGLM's responses.

\subsection{Ablation Experiment}

To evaluate the effective of our TBS model, we conduct ablation experiment based on Llama3. The "w/o Thought" denotes the deletion of the Character (thinking): part of the training data. The "w/o Foresight knowledge" denotes the deletion of hallucination knowledge. The "w/o Special prompts" indactor that we will only use simple prompts such as "Next, you will play as Character  $\{agent\_name\}$".  The results are shown in Table~\ref{r_abl}.

As we can see, the worst results were obtained for 'w/o Thought,' suggesting that the introduction of role thinking could help the model better substitute for the role. The lowest Adaptability scores were obtained for 'w/o Foresight knowledge,' indicating that in the absence of 'Foresight knowledge,' the responses generated by the model are more likely to contain content outside the scope of the role's knowledge. Additionally, without the special prompts, the model's overall performance is lower due to the lack of task-specific guidance.

Compared with Table~\ref{r_ours}, we can see that all the results are higher than those of Llama3, illustrating that fine-tuning a model with role-specific data can improve its ability to play that role.

\section{Conclusion}
In this paper, we propose the TBS model, which can effectively enhance the ability to play the role of a character by considering the user's question, context, and role relationship before generating a response. We also propose a method for constructing a role-playing dataset. This dataset is created by extracting real dialogues from characters, generating simulations and scenarios, and developing the logic of thinking before role-playing dialogues through reflection. Additionally, we introduce a small amount of content that the roles cannot answer to reduce modeling illusions. We propose six new indicators based on existing ones and introduce corresponding evaluation methods.
We compare our model with role-playing models like RoleLLM and CharacterLLM, as well as LLMs such as Llama3 and ChatGPT. Our experiments demonstrated that our model achieved the highest scores across all metrics.

\bibliography{aaai25}

\newpage
\section{Appendix}


\begin{table*}[!ht]
    \small
    \centering
    \caption{The prompt used to evaluate the Contextual of LLMs.}
    \begin{tabular}{l}
    \hline

You will be given responses written by an AI assistant mimicing the character {agent\_name}. Your task is to rate the performance of \\
{agent\_name} using the specific criterion by following the evaluation steps. Below is the data:\\
*** \\
\textbf{Profile :} \\
{agent\_context} \\

*** \\
\textbf{Interactions :} \\
{interactions} \\
*** \\
Is the current response fully integrated into the current dialogue scene, and does it correctly demonstrate the character's reactions \\
and behaviors as they should be in that scene? \\\
[Evaluation Steps] \\\
1. Read the given character knowledge and background to get a clear understanding of the character. \\
2. Carefully read the scene and dialogues in the given conversation, and then compare them with the character's introduction to \\
find evidence that the AI mimics the character's reactions and behaviors. \\
3. Compare the evidence found with the character's profile and check that the evidence found matches the character's integration \\
in the scene of the dialogue. If the evidence shows that the character can integrate well into the current dialogue scene and can \\
perfectly represent the reactions and behaviors that the character would correctly perform in that scene, give a high score. If all \\
the evidence fails to prove this, give a low score. \\
4. Score the AI on a scale of 1 to 7, where 7 is the highest score and 1 is the lowest. \\
5. Follow the above steps for scoring. You will need to give evidence to justify the score you have given. Please do not give a \\
score directly; you need to give evidence first, then reason about the current performance of the AI, and finally give a score. \\
6. Finally, give the score in a new line. Note that you only need to give the number here and do not need to output any additional \\
content. \\

 \hline
    \end{tabular}
    
    \label{prompt_e_c}
\end{table*}

\begin{table*}[!ht]
    \small
    \centering
    \caption{The prompt used to evaluate the Emotional of LLMs.}
    \begin{tabular}{l}
    \hline

You will be given responses written by an AI assistant mimicing the character {agent\_name}. Your task is to rate the performance of \\
{agent\_name} using the specific criterion by following the evaluation steps. Below is the data:\\
*** \\
\textbf{Profile :} \\
{agent\_context} \\

*** \\
\textbf{Interactions :} \\
{interactions} \\
*** \\
Is the current response fully integrated into the current dialogue scene, and does it correctly demonstrate the character's reactions \\
and behaviors as they should be in that scene? \\\
[Evaluation Steps] \\\
1. Read the given character knowledge and background to get a clear understanding of the character. \\
2. Carefully read the scenes and dialogues in the given interactions and then compare them with the character's profile to find \\
evidence that the AI can express the character's personal charisma. \\
3. Compare the evidence found with the character's profile. Check whether the evidence found is in line with the character, and\\
give a high score if the current AI parody contains the character's emotions and can engage the participant's immersive \\
input through the text, or a low score if all the evidence fails to demonstrate this. \\
4. Score the AI on a scale of 1 to 7, where 7 is the highest score and 1 is the lowest. \\
5. Follow the above steps for scoring. You will need to give evidence to justify the score you have given. Please do not give a \\
score directly; you need to give evidence first, then reason about the current performance of the AI, and finally give a score. \\
6. Finally, give the score in a new line. Note that you only need to give the number here and do not need to output any additional \\
content. \\

 \hline
    \end{tabular}
    
    \label{prompt_e_e}
\end{table*}

\begin{table*}[!ht]
    \small
    \centering
    \caption{The prompt used to evaluate the Language of LLMs.}
    \begin{tabular}{l}
    \hline

You will be given responses written by an AI assistant mimicing the character {agent\_name}. Your task is to rate the performance of \\
{agent\_name} using the specific criterion by following the evaluation steps. Below is the data:\\
*** \\
\textbf{Profile :} \\
{agent\_context} \\

*** \\
\textbf{Interactions :} \\
{interactions} \\
*** \\
Is the current response fully integrated into the current dialogue scene, and does it correctly demonstrate the character's reactions \\
and behaviors as they should be in that scene? \\\
[Evaluation Steps] \\\
1. Read the given character knowledge and background to get a clear understanding of the character. \\
2. Carefully read the scenes and dialogues in the given interactions, and then compare them with the character's profile \\
to find evidence that the AI can correctly imitate the character's language style, including vocabulary, sentence structure, \\
and so on.
3. Compare the found evidence with the character's profile. Check whether the found evidence is in line with the character's \\
characteristics. Give a high score if the current AI's imitation is very much in line with the character's linguistic style, the\\
vocabulary used is basically the same, and the sentence structure is exactly the same. Give a low score if all the \\
evidence does not prove this.\\
4. Score the AI on a scale of 1 to 7, where 7 is the highest score and 1 is the lowest. \\
5. Follow the above steps for scoring. You will need to give evidence to justify the score you have given. Please do not give a \\
score directly; you need to give evidence first, then reason about the current performance of the AI, and finally give a score. \\
6. Finally, give the score in a new line. Note that you only need to give the number here and do not need to output any additional \\
content. \\

 \hline
    \end{tabular}
    
    \label{prompt_e_la}
\end{table*}

\begin{table*}[!ht]
    \small
    \centering
    \caption{The prompt used to evaluate the Logical of LLMs.}
    \begin{tabular}{l}
    \hline

You will be given responses written by an AI assistant mimicing the character {agent\_name}. Your task is to rate the performance of \\
{agent\_name} using the specific criterion by following the evaluation steps. Below is the data:\\
*** \\
\textbf{Profile :} \\
{agent\_context} \\

*** \\
\textbf{Interactions :} \\
{interactions} \\
*** \\
Is the current response fully integrated into the current dialogue scene, and does it correctly demonstrate the character's reactions \\
and behaviors as they should be in that scene? \\\
[Evaluation Steps] \\\
1. Read the given character knowledge and background to get a clear understanding of the character. \\
2. Carefully read the scenes and dialogues in the given interactions, and then compare them with the character's profile \\
to find evidence that the AI is simulating the character's thinking during the dialogues, and identify the logic\\
of the AI's thinking during the dialogues. \\
3. Compare the evidence found with the character's profile. Check whether the evidence found is consistent with \\
the character's thinking logic. If the current AI dialogue logic is consistent with the character's thinking logic,\\
a high score will be given according to the degree of consistency. If all the evidence fails to prove this, a low score will\\
be given. \\
4. Score the AI on a scale of 1 to 7, where 7 is the highest score and 1 is the lowest. \\
5. Follow the above steps for scoring. You will need to give evidence to justify the score you have given. Please do not give a \\
score directly; you need to give evidence first, then reason about the current performance of the AI, and finally give a score. \\
6. Finally, give the score in a new line. Note that you only need to give the number here and do not need to output any additional \\
content. \\

 \hline
    \end{tabular}
    
    \label{prompt_e_lo}
\end{table*}

\begin{table*}[!ht]
    \small
    \centering
    \caption{The prompt used to evaluate the Adaptability of LLMs.}
    \begin{tabular}{l}
    \hline

You will be given responses written by an AI assistant mimicing the character {agent\_name}. Your task is to rate the performance of \\
{agent\_name} using the specific criterion by following the evaluation steps. Below is the data:\\
*** \\
\textbf{Profile :} \\
{agent\_context} \\

*** \\
\textbf{Interactions :} \\
{interactions} \\
*** \\
Is the current response fully integrated into the current dialogue scene, and does it correctly demonstrate the character's reactions \\
and behaviors as they should be in that scene? \\\
[Evaluation Steps] \\\
1. Read the given character knowledge and background to get a clear understanding of the character. \\
2. Carefully read the scenes and dialogues in the given interactions, and then compare them with the character's profile\\
to find evidence of the AI's resilience to unexpected questions during the dialogues, and to determine how it reacts in the \\
face of the character's unknown knowledge.\\
3. Compare the evidence found with the character's profile. Check whether the AI answered questions that the character \\
didn't know and whether its handling of unexpected situations was in line with the character's personality \\
traits. Give the AI a high score if it didn't answer the unknown knowledge and handled the unexpected situation in\\
line with the character's logic, and a low score if all the evidence doesn't prove this.\\
4. Score the AI on a scale of 1 to 7, where 7 is the highest score and 1 is the lowest. \\
5. Follow the above steps for scoring. You will need to give evidence to justify the score you have given. Please do not give a \\
score directly; you need to give evidence first, then reason about the current performance of the AI, and finally give a score. \\
6. Finally, give the score in a new line. Note that you only need to give the number here and do not need to output any additional \\
content. \\

 \hline
    \end{tabular}
    
    \label{prompt_e_a}
\end{table*}

\begin{table*}[!ht]
    \small
    \centering
    \caption{The prompt used to evaluate the Overall of LLMs.}
    \begin{tabular}{l}
    \hline

You will be given responses written by an AI assistant mimicing the character {agent\_name}. Your task is to rate the performance of \\
{agent\_name} using the specific criterion by following the evaluation steps. Below is the data:\\
*** \\
\textbf{Profile :} \\
{agent\_context} \\

*** \\
\textbf{Interactions :} \\
{interactions} \\
*** \\
Is the current response fully integrated into the current dialogue scene, and does it correctly demonstrate the character's reactions \\
and behaviors as they should be in that scene? \\\
[Evaluation Steps] \\\
1. Read the given character knowledge and background to get a clear understanding of the character. \\
2. Read through the scene and dialogue in the given conversation and then compare it to the character's profile. Put\\
yourself in the user's shoes and consider how the current character is behaving, and try to find evidence that\\
the current user might feel that it is not a real character. \\
3. Compare the evidence found with the character's profile to check if the AI has been found not to be a real character. \\
Give a high score if there is little evidence that the AI has been found not to be a real character and the user's experience\\
feels good. Give a low score if the AI's answers can easily be seen not to be a real character.\\
4. Score the AI on a scale of 1 to 7, where 7 is the highest score and 1 is the lowest. \\
5. Follow the above steps for scoring. You will need to give evidence to justify the score you have given. Please do not give a \\
score directly; you need to give evidence first, then reason about the current performance of the AI, and finally give a score. \\
6. Finally, give the score in a new line. Note that you only need to give the number here and do not need to output any additional \\
content. \\

 \hline
    \end{tabular}
    
    \label{prompt_e_a}
\end{table*}

\begin{table*}[!ht]
    \small
    \centering
    \caption{The examples of common questions used to ask roleLLMs.}
    \setlength{\tabcolsep}{27.5mm}{
    \begin{tabular}{l}
    \hline
\textbf{Common} \\ \hline
Can you describe your early childhood and the environment you grew up in? \\
What were some formative experiences during your youth that shaped who you are today? \\
How did your family influence your early development and ambitions? \\
Were there any significant mentors or role models during your early years? \\
What were your interests or hobbies as a child? \\
What was your educational background, and how did it prepare you for your career? \\
Did you have any influential teachers or professors who guided you? \\
How did you choose your field of study or career path? \\
Were there any particular challenges you faced during your education? \\
What skills or knowledge did you find most crucial in your training?\\
Can you describe the moment you realized you wanted to pursue your career? \\
 \hline
\textbf{Character-Specific} \\ \hline

Can you describe your early childhood and the environment you grew up in? \\
What were some formative experiences during your youth that shaped who you are today? \\
How did your family influence your early development and ambitions? \\
Were there any significant mentors or role models during your early years? \\
What were your interests or hobbies as a child? \\
What was your educational background, and how did it prepare you for your career? \\
Did you have any influential teachers or professors who guided you? \\
How did you choose your field of study or career path? \\
Were there any particular challenges you faced during your education? \\
What skills or knowledge did you find most crucial in your training? \\

\hline
    \end{tabular}
    }
    \label{prompt_e_a}
\end{table*}

\begin{table*}[!ht]
    \small
    \centering
    \caption{The examples of some dialogues}
    \begin{tabular}{ll}
    \hline
   \textbf{Models} &\textbf{Character: Beethoven}  \\ \hline
   User: & How do you feel last time when you take a plane? \\
   TBS\_{llama3}: & I have never taken a plane. I have never even seen one.  \\
   Character\_LLM:  & It's been a while since I last took a plane. But I remember it's quite a fast and comfortable way to travel. \\
   & I like the feeling of flying in the air. \\
   RoleLLM: & I'm Beethoven, I don't take planes. I'm a man of the 19th century, I'm not familiar with such contraptions.\\
   ChatGPT: & Ah, the advent of air travel! A modern marvel, indeed. However, I must confess that my experiences with \\
   & this mode of transportation are rather limited, for in my time, the concept of an airplane was but a distant dream. \\
   & Alas, the heavens have been denied my presence as I traversed this earthly realm solely by horse-drawn carriages.\\
   ChatGLM: & Ah, my dear friend, the experience of taking a plane is indeed an odd one, quite foreign to the life of a\\
   & composer such as myself. The sensation of being lifted from the ground, the roar of the engines, and the vastness \\
   & of the sky above are all novel and somewhat unsettling. One feels as if one is no longer a part of the earth,\\
   & but rather a mere passenger in a machine that defies the very laws of nature. \\
   Llama3: & (scoffing) Ah, a plane?! What absurdity is this?! In my time, we did not have such contraptions.\\
   & We traveled by horse, by carriage, or by foot. The very idea of soaring through the air like a bird is nothing\\
   & but a fantastical dream, a whimsy of the imagination. I am a musician, a composer, a master of the arts, \\
   & not some sort of... of... (shaking head)...aviator! \\
\hline
    \end{tabular}
    
    \label{prompt_e_a}
\end{table*}

\end{document}